\documentclass[letterpaper, 10 pt, conference]{ieeeconf}  
\usepackage[noadjust]{cite}                                                         
\usepackage{amsmath}
\usepackage{hyperref}

\usepackage{flushend} 

\usepackage{graphicx}
\usepackage{rotating}
\usepackage{array}
\usepackage{booktabs}
\newcommand{\head}[1]{\textnormal{\textbf{#1}}}

\IEEEoverridecommandlockouts                              
\title{\LARGE \bf
Automated Machine Learning in Practice: \\
State of the Art and Recent Results
}

\author{Lukas Tuggener$^{1,2}$, Mohammadreza Amirian$^{1,3}$, Katharina Rombach$^{1}$, Stefan L\"orwald$^{4}$,\\ Anastasia Varlet$^{4}$, Christian Westermann$^{4}$, and Thilo Stadelmann$^{1}$\\ \\
\parbox{5 cm}{$^{1}$ ZHAW Datalab, \\Winterthur, Switzerland\\ {\small \{tugg, amir, romc, stdm\}@zhaw.ch}} 
\parbox{6 cm}{$^{2}$ USI, Lugano, Switzerland\\ $^{3}$ Ulm University, Ulm, Germany \\}
\parbox{5 cm}{$^{4}$ PricewaterhouseCoopers AG (PwC), Zurich, Switzerland\\ {\small \{firstname.lastname\}@ch.pwc.com}}
}

\begin{document}

\maketitle
\thispagestyle{empty}
\pagestyle{empty}

\begin{abstract}
A main driver behind the digitization of industry and society is the belief that data-driven model building and decision making can contribute to higher degrees of automation and more informed decisions. Building such models from data often involves the application of some form of machine learning. Thus, there is an ever growing demand in work force with the necessary skill set to do so. This demand has given rise to a new research topic concerned with fitting machine learning models fully automatically---AutoML. This paper gives an overview of the state of the art in AutoML with a focus on practical applicability in a business context, and provides recent benchmark results on the most important AutoML algorithms.
\end{abstract}


\section{INTRODUCTION}
Many organisations, private and public, have understood that data analysis is a powerful tool to learn insights on how to improve their business model, decision making and even products \cite{braschler2019}. A central step in the analysis process often is the construction and training of a machine learning model, which itself entails several challenging steps, most notably feature preprocessing, algorithm selection, hyperparameter tuning and ensemble building. Usually a lot of expert knowledge is necessary to successfully perform all these steps \cite{meier2018learning}. The field of automated machine learning (AutoML) aims to develop methods that build suitable machine learning models without (or as little as possible) human intervention \cite{hutter2019automatic}. While there are many possible ways to state the AutoML Problem, we here focus on systems that address the "Combined Algorithm Selection and Hyperparameter optimization" (CASH) problem \cite{thornton2013auto}. A solver for the CASH problem aims to pick an algorithm from a list of options and then tune it to give the highest validation performance amongst all the (algorithm, hyperparameter) combinations.

In this paper, we give a comprehensive overview of the current state of AutoML and present new independent benchmark results of currently out of the box available systems as well as of cutting edge research. The next chapter gives insight why AutoML is currently not only heavily researched, but also of practical relevance in business and industry. Chapter \ref{chp:3} gives an overview of the current state of AutoML by introducing the most important concepts and systems. In chapter \ref{chp:4}, we present benchmark results between the scientific state of the art and an industrial prototype; we then discuss them with regard to practical AutoML system design choices in chapter \ref{chp:5}. Lastly, in chapter \ref{chp:6} we give a summary and outlook on approaches that will likely play a role the next important advancements in AutoML.

\section{IMPACT OF PRACTICAL AUTOMATED ML}
\label{chp:2}
In recent years, machine learning has been applied to more and more domains. Industrial applications for example, such as predictive maintenance \cite{susto2015machine,li2014improving} and defect detection \cite{figueiredo2011machine}, enable companies to be more proactive and improve efficiency. In the area of healthcare, patient data have helped addressing complex diseases, such as multiple sclerosis, and support doctors in identifying the most appropriate therapy \cite{phrend}. In the insurance and banking sectors, risks in 
loan applications \cite{handzic2003neural} and claims processing \cite{viaene2005auto,perez2005consolidated} can be estimated, enabling automated identification of fraudulent patterns. Finally, advances in sales and revenue forecasting support supply chain optimization \cite{tsoumakas2018survey}.

However, the process towards building such actionable machine learning models, able to generate added value to the business, is time-consuming and error-prone, if done manually; performance of different models should be compared, considering different algorithms, hyperparameter tuning and feature selection. This process is highly iterative and as such is a ideal candidate for automation. With the use of AutoML, the data scientist is freed from this tedious task and can focus on more creative tasks, delivering more value to the company. New business cases can be identified, assessed and validated in a rapid-prototype-fashion.

In practice, AutoML can provide different kind of insights. Already at an early stage, running different models on the input data can provide feedback on how suitable the data is for predicting the given target. When multiple models built from a wide spectrum of algorithms do not perform significantly better than the baseline, this can be seen as an indication of insufficient predictive power in the data. Ideally, however, good models will be generated and the data scientist is left with the choice of deploying the best generated model or to create an ensemble from a selection of models. Finally, there is a by-product in AutoML when optimizing over the feature set as well: One can derive an estimate of feature importance by statistical analysis of the model quality depending on which features are used as input to the models.

Thus, the introduction of AutoML tools in a company can drastically increase efficiency of the work of a data scientist. Taking the example of one of our predictive maintenance projects in the area of public transport, where a team of three data scientists worked full time for weeks, a machine learning model with an out-of-sample area under the curve (AUC) of $0.81$ was developed. A few months after that milestone, the prototype of the AutoML tool described later in Section \ref{chp:4} (DSM), using the same dataset (and no other help or information) was used as a benchmark to evaluate the benefits of this tool. It resulted in an automatically generated model that was slightly out-performing the manually engineered model with an AUC of $0.82$ after a run time of half an hour.

\section{STATE OF THE ART IN AUTOMATED ML}
\label{chp:3}
The problem of manual hyperparameter tuning \cite{li2017hyperband} inspired researchers to automate various blocks of the machine learning pipeline: \emph{feature engineering} \cite{duan2018automated}, \emph{meta-learning} \cite{andrychowicz2016learning}, \emph{architecture search} \cite{zoph2016neural} as well as full \emph{Combined Model Selection and Hyperparameter optimization} \cite{feurer2015efficient} are the research lines which grabbed a great deal of attention in the past years. We review them in this order.

\textbf{Feature engineering}: Feature preprocessing, representation learning and selecting the most discriminant features for a given classification or regression task are problems targeted by the literature. Gaudel et al. \cite{gaudel2010feature} consider feature engineering as a one-player game and train a reinforcement learning-based agent to select the best features. To do so, they first model the feature selection problem as a Markov Decision Process (MDP). Second, they propose a reward associated with generalization error in the final status. 
The agent learns an optimal policy to minimize the final generalization error.

\emph{Explorekit} \cite{katz2016explorekit} not only iteratively selects the features but also generates new feature candidates to obtain the most discriminant ones. Katz et al. use normalization and discrimination operators on a single feature to generate unary features. They additionally combine two or more features to generate new candidates and train a feature rank estimator based on meta-features extracted from the datasets and candidates.
The feature with the highest rank that increases the classification accuracy above a certain threshold is added to the selected feature set in every iteration.

To reduce the computational complexity of iterative feature selection methods, \emph{Learning Feature Engineering} (LFE) \cite{nargesian2017learning} learns the effectiveness of a transformation based on previous experiments. The original feature space is subsequently mapped via the optimal transformation to compute a discriminant feature representation.

\emph{AutoLearn} \cite{kaul2017autolearn} is a regression-based algorithm for automatic feature selection. The proposed algorithm starts by filtering the original features and discard the ones with small information gain. Subsequently, feature pairs are filtered based on distance correlation to omit dependent pairs. The new features are generated based on the remaining pairs using ridge regression. Ultimately, the best features are the ones with the highest information gain and stability \cite{meinshausen2010stability}. AutoLearn has been applied to a range of datasets including Gene expression data.

\textbf{Meta-learning} here refers to methods that try to leverage meta information about the problem at hand, e.g. the dataset as well as the available algorithms and their configurations, to improve the performance of an AutoML system. This meta-information is often gathered and processed using machine learning methods, thus in a sense applying the discipline to itself. The meta information about datasets often consists of some basic statistical reference numbers and some landmarks, i.e. performance figures of simple algorithms \cite{pfahringer2000meta}. 

\emph{Learning curve prediction} attempts to learn a model that predicts how much the performance of a learner will improve if given more training time \cite{klein2016learning}. Another take on this idea are attempts to predict the running time of algorithms \cite{eggensperger2018neural}. Instead of predicting absolute performance figures, it has sometimes proven beneficial to predict a ranking of the available algorithms to choose from \cite{brazdil2000comparison}.

\emph{Meta-learners} in the context of neural networks aim to improve the optimizer of a deep or shallow (convolutional) neural network (CNN) to reach a minimum as quick as possible through automatic hyper-parameter tuning. Andrychowicz et al. \cite{andrychowicz2016learning} learn to predict the best set of hyperparameters for optimizing neural networks with gradients and a Long Short-Term Memory \cite{hochreiter1997long} network. Similarly, Chen et al. \cite{chen2017learning} train an optimizer for simple synthetic functions such as Gaussian Processes. They demonstrate that the optimizer generalizes to a wide range of black-box problems. For instance, the trained optimizer is used to tune the hyperparameters of a Support Vector Machine \cite{cortes1995support} without accessing the gradients of the loss function with respect to the hyperparameters.  

\textbf{Architecture search} literature discusses methods for finding the best performing architecture for neural networks automatically without human expert intervention. Elsken et al. \cite{elsken2017simple} propose \emph{Neural Architecture Search by Hill-climbing} (NASH) using local search. The algorithm starts with a well-performing (preferably pretrained) convolutional architecture (parent). Then, two types of network morphisms (transformations) are randomly applied to generate deeper or wider architecture children from the original parent network. The children architectures are trained, and the best-performing architecture qualifies for the next round. The algorithm iterates until the validation accuracy saturates.    

Real et al. \cite{pmlr-v70-real17a} propose an evolutionary architecture search based on a pairwise comparison within the population: the algorithm starts with an initial population as parents, and every network undergoes random mutations such as adding and removing convolutional layers and skip connections to produce offspring. Subsequently, parent and child compete in pairwise comparison, with the winner model staying in the population and the loser being discarded.

In contrast to evolutionary algorithms, where larger and more accurate architectures are desired, He et al. \cite{he2018amc} automatically search for compressing a given CNN for mobile and embedded applications. Their \emph{AutoML for Model Compression} (AMC) algorithm trains a reinforcement learning agent to estimate the sparsity ratio of each layer and then compress the layers sequentially.

Zoph et al. \cite{zoph2016neural} train a controller using reinforcement learning and a Recurrent Neural Network (RNN) to tune the hyperparameters of a deep CNN architecture such as width and height of filters and strides. The RNN controller is trained using a policy gradient method to maximize the network's accuracy on a hold-out set of data. 

\textbf{CASH}: The main focus of this paper lies on the \emph{Combined Model Selection and Hyperparameter optimization} problem, which can be solved by employing a combination of building blocks mentioned above. 
Ultimately, a full solution finds the best machine learning pipeline for raw (un-preprocessed) feature vectors in the shortest time for a given amount of computational resources. This inspired the series of Automated Machine Learning (AutoML) challenges since 2015 \cite{guyon2017analysis}. A complete pipeline includes data cleaning, feature engineering (selection and construction), model selection, hyperparameter optimization and finally building an ensemble of the top trained models to obtain good performance on unseen test data. Optimizing the entire machine learning pipeline (that is not necessarily differentiable end-to-end) is a challenging task, and different solutions have been investigated using various approaches.

\textbf{Hyperparameter optimization} is a crucial step for solving the entire CASH problem, with \emph{Bayesian optimization} \cite{brochu2010tutorial} being the most prominent specimen of respective approaches. The goal here is to build a model of expected loss and variance for every input. After each optimization step, the model (or current belief) is updated using the a posteriori information (hence the name Bayesian).

An acquisition function is defined that decides at which location to sample the next true loss, trading off regions of low expected loss (exploitation) and regions of high variance (exploration). Usually, Gaussian Processes are the model of choice in Bayesian optimization; alternatively, Random Forests have been used to model the loss surface of the hyperparameters as a Gaussian distribution in \emph{Sequential Model-based optimization for general Algorithm Configuration} (SMAC) \cite{hutter2011sequential} or the \emph{Tree-structured Parzen Estimator} \cite{feurer2014using}.

Model free methods include \emph{Successive Halving} \cite{jamieson2016non} and built on it \emph{Hyperband} \cite{li2017hyperband}, which uses real time optimization progress to narrow down a set of competing hyperparameter configurations over the duration of a full optimization run, possibly with many restarts. A slight variation of this are \emph{Evolutionary Strategies} that also allow for perturbations of the individual configurations during training \cite{jaderberg2017population}. In the special case where the optimizee as well as the optimizer are differentiable, multiple iterations of the optimizer can be unrolled, and an update for the hyperparamaters can be computed by using gradient descent and backpropagation \cite{maclaurin2015gradient}.

\textbf{Pipeline Optimization}: Complete machine learning pipelines, including feature prepossessing, model selection, hyperparameters optimization and building ensembles, are constructed based on different views of the entire problem. Bayesian optimization, Genetic programming \cite{banzhaf1998genetic}, and bandit optimization inspired developing various pipeline optimization frameworks.

\emph{Auto-sklearn} is aimed at solving the CASH problem using meta-learning, Bayesian optimization and ensemble building. First, it extracts meta-features of a new dataset such as task type (classification or regression), number of classes, the dimensionality of the feature vectors, number of samples and so on. The meta-learner of Auto-sklearn uses these meta-features to initialize the optimization step based on previous experience on similar data sets (similar according to the meta features). Then, preprocessing and model hyperparameters are iteratively enhanced using Bayesian optimization. Ultimately, a robust classifier or regression model is built based on an ensemble of models trained during the iterative optimization.

\emph{Tree-based Pipeline Optimization Tool} (\emph{TPOT}) is an algorithm uses genetic programming to optimize the machine learning pipeline. It searches for the best pipeline including feature processing, model and hyperparameters for a given classification or regression task. The feature processing module of TPOT works in conjunction with feature selection and generation. The feature generation block performs the kernel trick \cite{scholkopf2001kernel} or dimensionality reduction methods. The optimization is done using genetic programming: first, the algorithm generates some tree-based pipelines randomly. Then, it selects the top $20\%$ of the generated population based on cross-validation accuracy, and produces $5$ descendants from each by randomly changing a point in the pipeline. The algorithm continues until a stopping criterion is met.  

The \emph{ATM} framework \cite{swearingen2017atm} finally uses multi-armed bandit optimization in combination with hybrid Bayesian optimization to find the best models.

\label{Sec:State-of-art}
\section{EXPERIMENTAL EVALUATION}
\label{chp:4}
In this section, we evaluate the usefulness of AutoML for application in business and industry by empirically comparing the most successful automated machine learning algorithms with (a) an industrial prototype as well as (b) a straight-forward improvement inspired by Hyperband \cite{li2017hyperband, feurer2018practical} (c). This selection spans a wide range of different approaches for pipeline optimization (see Section \ref{Sec:State-of-art}) to tackle the CASH problem: the industrial prototype \emph{DSM} \cite{stadelmann2018deep} 
uses random model and hyperparameter search and thus serves as a baseline; Auto-sklearn \cite{feurer2015efficient} has won the recent AutoML challenges \cite{guyon2017analysis}. Additionally, we report results with TPOT \cite{olson2016automating}, which is developed based on genetic programming \cite{banzhaf1998genetic} instead of Auto-sklearn's Bayesian optimization. We gave a brief overview of each system below:

\textbf{Data Science Machine (DSM)}: The Data Science Machine (DSM) \cite{stadelmann2018deep} has been developed for both in-house and client-related data science projects by PwC. 
DSM includes a portfolio of open-source machine learning algorithms, and also offers the possibility to add custom algorithms through a language-agnostic API. Given a dataset, the tool automatically optimizes over the set of algorithms, features, and hyperparameters. The developed solution offers multiple optimization strategies, e.g. tuneable genetic algorithms. In the context of this paper, however, we limited it to use random sampling to serve as a baseline.

\textbf{Auto-sklearn} \cite{feurer2015efficient}: We used the algorithm explained in Section \ref{Sec:State-of-art} to find the best pipeline within the time-budget computed for training $100$ models by DSM. Auto-sklearn is slow in start \cite{feurer2015efficient}; however, it eventually reaches a solution very close to the optimum. The method benefits from meta-learning using similar datasets and it is usually pretrained on OpenML datasets \cite{OpenML2013} in the challenges. We applied the base model of Auto-sklearn to compare the exploration efficiency of the different approaches.           

\textbf{TPOT} \cite{olson2016automating}: This algorithm uses tree-based classifiers which is similar to the second entry of the latest AutoML challenge \cite{guyon2017analysis}. TPOT differs from the other presented methods since it used Genetic programming for optimization. We initialized the algorithm with a population of $20$ tree-based pipelines and stopped the optimization with the same time budget as DSM and Auto-sklearn.

\begin{figure}[h]
\includegraphics[width=0.45\textwidth]{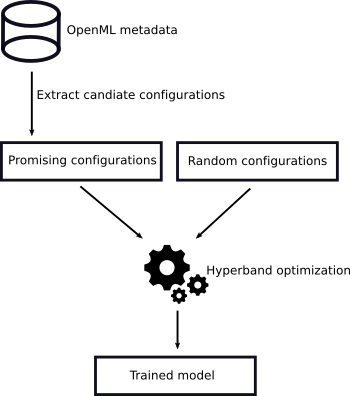}
\caption{Schematic overview of the \emph{Portfolio Hyperband} workflow.}
\end{figure}

\textbf{Portfolio Hyperband} \cite{feurer2018practical,li2017hyperband}: Inspired by PoSH Auto-sklearn
\cite{feurer2018practical} that combines a portfolio of initial configurations with successive halving (SH) and Bayesian optimization, we tested a system that combines a portfolio with Hyperband \cite{li2017hyperband}. 
Our goal was to combine the portfolio variant of meta-learning, which is very simple and fast, with Hyperband that should give the system a good asymptotic performance.
At the basis of Hyperband is the successive halving algorithm that starts with an initial set of configurations and trains all of them for a fixed amount of time; then, the less performing half of the configurations is dropped. This process is repeated until only the best performing configuration remains, hence the name successive halving. An issue with SH is that it is unclear with how many initial configurations to start the process. Hyperband performs a geometric search to explore the trade-off between individual training time and the amount of different initial configurations.

In order to create a collection of promising configuration candidates, we surveyed all the meta data available on OpenML \cite{OpenML2013} and extracted a list of different configurations that worked well for binary classification. This list is used to seed Hyperband. Every run also uses some random initialisations to improve long term performance. To test the viability of Portfolio Hyperband we applied it to binary classification only.

To compare the presented automated machine learning approaches, we select some datasets with various classification and regression tasks from previous AutoML challenges \cite{guyon2017analysis}. We chose all datasets that are supported by DSM, TPOT and Auto-sklearn. Because the official test labels are not public we used our own training, validation, test split. For the DSM (random search as a baseline), we randomly pick a set of $100$ models and hyperparameters, train the models and extract the best-performing ones for each data set. During the experiments, the time budget used for DSM is computed and used as the reference value for the rest of the optimization methods. Auto-sklearn, TPOT and Portfolio Hyperband subsequently use this time budget to find the best pipeline. We start training from scratch without pretraining on any similar data set; therefore, the meta-learning block of Auto-sklearn is not well tuned. Since Portfolio Hyperband turned out to find good initial models very fast, we also report its performance after a considerably shortened period of 10 minutes.

\begin{table*}[t!]
	\centering
	\resizebox{\textwidth}{!}{\begin{tabular}{l l l|  c c c c c c|cc}
        \toprule[1.5pt]
        & & &  \multicolumn{2}{c}{\head{DSM}} & \multicolumn{2}{c}{\head{Auto-Sklearn}} &
        \multicolumn{2}{c}{\head{TPOT}} & \multicolumn{2}{c}{\head{Portfolio Hyperband}} \\
        \textbf{Dataset} & \textbf{Task} & \textbf{Metric}  & \textbf{Test} & \textbf{Time}  & \textbf{Test} & \textbf{Time}  &  \textbf{Test} & \textbf{Time}  & \textbf{Test, full time} & \textbf{Test, 10 Min} \\
        \cmidrule(lr){1-11}

Cadata        & Regression                   & R2 (coefficient of determination)    &        0.7119        &      55.0    &       0.7327        &      54.9       &     \textbf{0.7989} &      54.6              & - & - \\
Christine     & Binary classification        & Balanced accuracy score              &        0.7146        &      99.4    &       0.7392        &      99.3       &       0.7442        &      105.1           & \textbf{0.753} & \textbf{0.744}  \\
Digits        & Multiclass classification    & Balanced accuracy score              &        0.8751        &      201.2   &     \textbf{0.9542} &      201.2      &       0.9476        &      207.2       & - & -   \\
Fabert        & Multiclass classification    & Accuracy score                       &        0.8665        &      77.5    &     \textbf{0.8908} &      77.4       &       0.8835        &      78.5       & - & -    \\
Helena        & Multiclass classification    & Balanced accuracy score              &        0.2103        &      190.2   &       0.3235        &      216.4      &     \textbf{0.3470} &      197.5       & - & -   \\
Jasmine       & Binary classification        & Balanced accuracy score              &     \textbf{0.8371}  &      24.1    &       0.8214        &      24.0       &       0.8326        &      25.9         & - & -   \\
Madeline      & Binary classification        & Balanced accuracy score              &        0.7686        &      48.3    &     \textbf{0.8896} &      48.2       &       0.8684        &      53.0        & 0.868 & 0.848  \\
Philippine    & Binary classification        & Balanced accuracy score              &        0.7406        &      56.3    &       0.7634        &      56.2       &     \textbf{0.7703} &      56.4       & 0.741 & 0.753  \\
Sylvine       & Binary classification        & Balanced accuracy score              &        0.9233        &      28.9    &       0.9350        &      28.9       &     \textbf{0.9415} &      29.0         & \textbf{0.947} & 0.916  \\
Volkert       & Multiclass classification    & Accuracy score                       &        0.8154        &      122.3   &     \textbf{0.8880} &      122.2      &       0.8720        &      125.5        & - & -  \\
\midrule
\multicolumn{3}{l|}{\head{Average Performance}}                                     &        0.7463        &\textbf{90.31} &      0.7938        &      92.85      &     \textbf{0.8006} &      93.26        &   &    \\
        \bottomrule[1.5pt]
    \end{tabular}}
    \caption{Performance of selected automated machine learning algorithms on AutoML challenge data sets \cite{guyon2017analysis}. "Test" refers to our own test split; "Time" is in minutes; "Portfolio Hyperband" only tested on binary classification.}
	\label{table:1}
\end{table*}

	


The results of the benchmarking are presented in Table \ref{table:1}. Numerical evaluation suggests that all three sophisticated approaches outperform the DSM in baseline mode (only random search) consistently, but not by a large margin. 

However, the accuracy of the three approaches from current research is quite similar, while Portfolio Hyperband appears to be especially quick. 

\section{DISCUSSION}
\label{chp:5}
As we showed in the previous section, none of the presented methods is clearly superior to all of the others. We can identify two major takeaways. First, we note that random search (DSM) is still quite competitive, especially when it is constrained to a relatively small set of tried and true options. It falls short with respect to systems that leverage meta-learning, especially with very constrained time budgets. A take away from this is that sometimes it can make sense to invest in faster and more hardware for parallel search rather than in a very sophisticated AutoML solution. 

Second, we find that the use of meta data to guide the search or even pre-trained models is one of the most potent ways to speed up AutoML. Working with completely unrelated data (e.g. from OpenML) already yields a sizeable speed-up in the Portfolio Hyperband system. 
The closer the data on which the meta-learner is trained (offline data) and the data to be analyzed (online data) are in distribution, the the stronger this effect gets---up to the point where the model can be almost fully trained on the offline data and then just fine tuned using the online data. Meta learning therefore is especially attractive for business cases where a continuous data generating process (e.g. monthly reports, continuous sensor feedback) produces new data that is similar in distribution to the old data already seen from the same source.


\section{SUMMARY AND OUTLOOK}
\label{chp:6}
As the amount of digital data is rapidly growing, data-driven approaches such as shallow and deep machine learning are increasingly used, in turn increasing the demand for efficient and generalizable AutoML. In this paper, we have presented an independent evaluation of current approaches in the field of shallow AutoML, and have tested our version of Portfolio Hyperband that shows promising results in terms of computational efficiency while being on par with the state of the art accuracy-wise.

Current AutoML approaches can be summarized as carefully engineered systems based on a collection of well established ideas. In this context, the use of meta-information for efficiency reasons---i.e., making the best of the used compute time---seems to be the most expandable idea. It bears similarity to the way human machine learning experts use their experience to solve a machine learning task, and we have surveyed some very successful attempts to utilize previously trained meta-knowledge in AutoML setups above \cite{feurer2015efficient}. Yet, the effectiveness of the meta-information
is highly dependent on the similarity of the tasks at hand to the ones that have been encountered in the meta-learning step, as well as on the metric to determine this similarity.

Future improvements in the field will thus likely be made based on more general concepts instead of just further engineering. Of special interest in this regard is one line of meta-learning research that aims at learning to exploit the intrinsic structure in the problems of interest in an automatic way \cite{andrychowicz2016learning}. Such a  meta-learner is trained on a variety of objective functions (directly or indirectly formulated) and learns how to efficiently move on the objective function's response surface in order to find an optimum. In other words, the efficient exploration of high dimensional spaces is learned. 

This concept of \emph{learning to optimize} is currently lacking in the AutoML frameworks introduced in Section \ref{chp:3}. However, the approach has shown to be able to match the performance of heavily engineered Bayesian optimization solutions such as Spearmint, SMAC and TPE while being massively faster \cite{chen2017learning}. Thus, we consider the concept of learning to optimize a promising direction for future work. As it delivers a very general black box optimizer, it is well-suited to be applied to both AutoML as described in this paper as well as to architecture search in deep learning---an increasingly more relevant research topic. 

Current approaches of learning to optimize require the accessibility of the objective function's gradient in either the meta-training phase or both, the meta-training phase and during execution \cite{andrychowicz2016learning} \cite{chen2017learning} \cite{li2016learning}. As AutoML is a task of optimizing a non-smooth and not explicitly known objective function, there is often no accessibility to this gradient. Thus, future work aims at the idea of learning to optimize, but the meta-training paradigm is changed to reinforcement learning to enable training on more realistic, i.e.\ non-smooth objective functions.

\section*{Acknowledgement} We are grateful for support by Innosuisse grant 25948.1 PFES ``Ada'' and helpful discussions with Martin Jaggi.

\bibliographystyle{ieeetr}

\begin{thebibliography}{10}

\bibitem{braschler2019}
M.~Braschler, K.~Stockinger, and T.~Stadelmann~(Eds.), {\em Applied Data
  Science---Lessons Learned for the Data-Driven Business}.
\newblock Springer International Publishing, 2019.

\bibitem{meier2018learning}
B.~B. Meier, I.~Elezi, M.~Amirian, O.~D{\"u}rr, and T.~Stadelmann, ``Learning
  neural models for end-to-end clustering,'' in {\em IAPR Workshop on
  Artificial Neural Networks in Pattern Recognition}, pp.~126--138, Springer,
  2018.

\bibitem{hutter2019automatic}
F.~Hutter, L.~Kotthoff, and J.~Vanschoren, ``Automatic machine learning:
  methods, systems, challenges,'' {\em Challenges in Machine Learning}, 2019.

\bibitem{thornton2013auto}
C.~Thornton, F.~Hutter, H.~H. Hoos, and K.~Leyton-Brown, ``Auto-weka: Combined
  selection and hyperparameter optimization of classification algorithms,'' in
  {\em Proceedings of the 19th ACM SIGKDD international conference on Knowledge
  discovery and data mining}, pp.~847--855, ACM, 2013.

\bibitem{susto2015machine}
G.~A. Susto, A.~Schirru, S.~Pampuri, S.~McLoone, and A.~Beghi, ``Machine
  learning for predictive maintenance: A multiple classifier approach,'' {\em
  IEEE Transactions on Industrial Informatics}, vol.~11, no.~3, pp.~812--820,
  2015.

\bibitem{li2014improving}
H.~Li, D.~Parikh, Q.~He, B.~Qian, Z.~Li, D.~Fang, and A.~Hampapur, ``Improving
  rail network velocity: A machine learning approach to predictive
  maintenance,'' {\em Transportation Research Part C: Emerging Technologies},
  vol.~45, pp.~17--26, 2014.

\bibitem{figueiredo2011machine}
E.~Figueiredo, G.~Park, C.~R. Farrar, K.~Worden, and J.~Figueiras, ``Machine
  learning algorithms for damage detection under operational and environmental
  variability,'' {\em Structural Health Monitoring}, vol.~10, no.~6,
  pp.~559--572, 2011.

\bibitem{phrend}
E.~St\"uhler, S.~Braune, F.~Lionetto, Y.~Heer, P.~Kassraian-Fard, E.~Jules,
  C.~Westermann, A.~Bergmann, P.~van Hövell, and N.~S. Group, ``Framework for
  personalized prediction of treatment response in relapsing remitting multiple
  sclerosis,'' {\em BMC medical research methodology}, submitted.

\bibitem{handzic2003neural}
M.~Handzic, F.~Tjandrawibawa, and J.~Yeo, ``How neural networks can help loan
  officers to make better informed application decisions,'' {\em Informing
  Science}, vol.~6, pp.~97--109, 2003.

\bibitem{viaene2005auto}
S.~Viaene, G.~Dedene, and R.~A. Derrig, ``Auto claim fraud detection using
  bayesian learning neural networks,'' {\em Expert Systems with Applications},
  vol.~29, no.~3, pp.~653--666, 2005.

\bibitem{perez2005consolidated}
J.~M. P{\'e}rez, J.~Muguerza, O.~Arbelaitz, I.~Gurrutxaga, and J.~I.
  Mart{\'\i}n, ``Consolidated tree classifier learning in a car insurance fraud
  detection domain with class imbalance,'' in {\em International Conference on
  Pattern Recognition and Image Analysis}, pp.~381--389, Springer, 2005.

\bibitem{tsoumakas2018survey}
G.~Tsoumakas, ``A survey of machine learning techniques for food sales
  prediction,'' {\em Artificial Intelligence Review}, pp.~1--7, 2018.

\bibitem{li2017hyperband}
L.~Li, K.~Jamieson, G.~DeSalvo, A.~Rostamizadeh, and A.~Talwalkar, ``Hyperband:
  A novel bandit-based approach to hyperparameter optimization,'' {\em The
  Journal of Machine Learning Research}, vol.~18, no.~1, pp.~6765--6816, 2017.

\bibitem{duan2018automated}
J.~Duan, Z.~Zeng, A.~Oprea, and S.~Vasudevan, ``Automated generation and
  selection of interpretable features for enterprise security,'' in {\em 2018
  IEEE International Conference on Big Data (Big Data)}, pp.~1258--1265, IEEE,
  2018.

\bibitem{andrychowicz2016learning}
M.~Andrychowicz, M.~Denil, S.~Gomez, M.~W. Hoffman, D.~Pfau, T.~Schaul,
  B.~Shillingford, and N.~De~Freitas, ``Learning to learn by gradient descent
  by gradient descent,'' in {\em Advances in Neural Information Processing
  Systems}, pp.~3981--3989, 2016.

\bibitem{zoph2016neural}
B.~Zoph and Q.~V. Le, ``Neural architecture search with reinforcement
  learning,'' in {\em Proceedings of International Conference on Learning
  Representations (ICLR)}, 2017.

\bibitem{feurer2015efficient}
M.~Feurer, A.~Klein, K.~Eggensperger, J.~Springenberg, M.~Blum, and F.~Hutter,
  ``Efficient and robust automated machine learning,'' in {\em Advances in
  Neural Information Processing Systems}, pp.~2962--2970, 2015.

\bibitem{gaudel2010feature}
R.~Gaudel and M.~Sebag, ``Feature selection as a one-player game,'' in {\em
  International Conference on Machine Learning}, pp.~359--366, 2010.

\bibitem{katz2016explorekit}
G.~Katz, E.~C.~R. Shin, and D.~Song, ``Explorekit: Automatic feature generation
  and selection,'' in {\em Data Mining (ICDM), 2016 IEEE 16th International
  Conference on}, pp.~979--984, IEEE, 2016.

\bibitem{nargesian2017learning}
F.~Nargesian, H.~Samulowitz, U.~Khurana, E.~B. Khalil, and D.~Turaga,
  ``Learning feature engineering for classification,'' in {\em Proceedings of
  the Twenty-Sixth International Joint Conference on Artificial Intelligence,
  IJCAI}, vol.~17, pp.~2529--2535, 2017.

\bibitem{kaul2017autolearn}
A.~Kaul, S.~Maheshwary, and V.~Pudi, ``Autolearn—automated feature generation
  and selection,'' in {\em Data Mining (ICDM), 2017 IEEE International
  Conference on}, pp.~217--226, IEEE, 2017.

\bibitem{meinshausen2010stability}
N.~Meinshausen and P.~B{\"u}hlmann, ``Stability selection,'' {\em Journal of
  the Royal Statistical Society: Series B (Statistical Methodology)}, vol.~72,
  no.~4, pp.~417--473, 2010.

\bibitem{pfahringer2000meta}
B.~Pfahringer, H.~Bensusan, and C.~G. Giraud-Carrier, ``Meta-learning by
  landmarking various learning algorithms.,'' in {\em ICML}, pp.~743--750,
  2000.

\bibitem{klein2016learning}
A.~Klein, S.~Falkner, J.~T. Springenberg, and F.~Hutter, ``Learning curve
  prediction with bayesian neural networks,'' 2016.

\bibitem{eggensperger2018neural}
K.~Eggensperger, M.~Lindauer, and F.~Hutter, ``Neural networks for predicting
  algorithm runtime distributions.,'' in {\em IJCAI}, pp.~1442--1448, 2018.

\bibitem{brazdil2000comparison}
P.~B. Brazdil and C.~Soares, ``A comparison of ranking methods for
  classification algorithm selection,'' in {\em European conference on machine
  learning}, pp.~63--75, Springer, 2000.

\bibitem{hochreiter1997long}
S.~Hochreiter and J.~Schmidhuber, ``Long short-term memory,'' {\em Neural
  computation}, vol.~9, no.~8, pp.~1735--1780, 1997.

\bibitem{chen2017learning}
Y.~Chen, M.~W. Hoffman, S.~G. Colmenarejo, M.~Denil, T.~P. Lillicrap,
  M.~Botvinick, and N.~de~Freitas, ``Learning to learn without gradient descent
  by gradient descent,'' in {\em Proceedings of the 34th International
  Conference on Machine Learning-Volume 70}, pp.~748--756, JMLR. org, 2017.

\bibitem{cortes1995support}
C.~Cortes and V.~Vapnik, ``Support-vector networks,'' {\em Machine learning},
  vol.~20, no.~3, pp.~273--297, 1995.

\bibitem{elsken2017simple}
T.~Elsken, J.-H. Metzen, and F.~Hutter, ``Simple and efficient architecture
  search for convolutional neural networks,'' in {\em Proceedings of
  International Conference on Learning Representations (ICLR)}, 2018.

\bibitem{pmlr-v70-real17a}
E.~Real, S.~Moore, A.~Selle, S.~Saxena, Y.~L. Suematsu, J.~Tan, Q.~V. Le, and
  A.~Kurakin, ``Large-scale evolution of image classifiers,'' in {\em
  Proceedings of the 34th International Conference on Machine Learning}
  (D.~Precup and Y.~W. Teh, eds.), vol.~70 of {\em Proceedings of Machine
  Learning Research}, (International Convention Centre, Sydney, Australia),
  pp.~2902--2911, PMLR, 06--11 Aug 2017.

\bibitem{he2018amc}
Y.~He, J.~Lin, Z.~Liu, H.~Wang, L.-J. Li, and S.~Han, ``Amc: Automl for model
  compression and acceleration on mobile devices,'' in {\em Proceedings of the
  European Conference on Computer Vision (ECCV)}, pp.~784--800, 2018.

\bibitem{guyon2017analysis}
I.~Guyon, L.~Sun-Hosoya, M.~Boull{\'e}, H.~Escalante, S.~Escalera, Z.~Liu,
  D.~Jajetic, B.~Ray, M.~Saeed, M.~Sebag, {\em et~al.}, ``Analysis of the
  automl challenge series 2015-2018,'' 2017.

\bibitem{brochu2010tutorial}
E.~Brochu, V.~M. Cora, and N.~De~Freitas, ``A tutorial on bayesian optimization
  of expensive cost functions, with application to active user modeling and
  hierarchical reinforcement learning,'' {\em arXiv preprint arXiv:1012.2599},
  2010.

\bibitem{hutter2011sequential}
F.~Hutter, H.~H. Hoos, and K.~Leyton-Brown, ``Sequential model-based
  optimization for general algorithm configuration,'' in {\em International
  Conference on Learning and Intelligent Optimization}, pp.~507--523, Springer,
  2011.

\bibitem{feurer2014using}
M.~Feurer, J.~T. Springenberg, and F.~Hutter, ``Using meta-learning to
  initialize bayesian optimization of hyperparameters,'' in {\em Proceedings of
  the 2014 International Conference on Meta-learning and Algorithm
  Selection-Volume 1201}, pp.~3--10, Citeseer, 2014.

\bibitem{jamieson2016non}
K.~Jamieson and A.~Talwalkar, ``Non-stochastic best arm identification and
  hyperparameter optimization,'' in {\em Artificial Intelligence and
  Statistics}, pp.~240--248, 2016.

\bibitem{jaderberg2017population}
M.~Jaderberg, V.~Dalibard, S.~Osindero, W.~M. Czarnecki, J.~Donahue, A.~Razavi,
  O.~Vinyals, T.~Green, I.~Dunning, K.~Simonyan, {\em et~al.}, ``Population
  based training of neural networks,'' {\em arXiv preprint arXiv:1711.09846},
  2017.

\bibitem{maclaurin2015gradient}
D.~Maclaurin, D.~Duvenaud, and R.~Adams, ``Gradient-based hyperparameter
  optimization through reversible learning,'' in {\em International Conference
  on Machine Learning}, pp.~2113--2122, 2015.

\bibitem{banzhaf1998genetic}
W.~Banzhaf, P.~Nordin, R.~E. Keller, and F.~D. Francone, {\em Genetic
  programming: an introduction}, vol.~1.
\newblock Morgan Kaufmann San Francisco, 1998.

\bibitem{scholkopf2001kernel}
B.~Sch{\"o}lkopf, ``The kernel trick for distances,'' in {\em Advances in
  neural information processing systems}, pp.~301--307, 2001.

\bibitem{swearingen2017atm}
T.~Swearingen, W.~Drevo, B.~Cyphers, A.~Cuesta-Infante, A.~Ross, and
  K.~Veeramachaneni, ``Atm: A distributed, collaborative, scalable system for
  automated machine learning,'' in {\em IEEE International Conference on Big
  Data}, 2017.

\bibitem{feurer2018practical}
M.~Feurer, K.~Eggensperger, S.~Falkner, M.~Lindauer, and F.~Hutter, ``Practical
  automated machine learning for the automl challenge 2018,'' in {\em
  International Workshop on Automatic Machine Learning at ICML}, 2018.

\bibitem{stadelmann2018deep}
T.~Stadelmann, M.~Amirian, I.~Arabaci, M.~Arnold, G.~F. Duivesteijn, I.~Elezi,
  M.~Geiger, S.~L{\"o}rwald, B.~B. Meier, K.~Rombach, {\em et~al.}, ``Deep
  learning in the wild,'' in {\em IAPR Workshop on Artificial Neural Networks
  in Pattern Recognition}, pp.~17--38, Springer, 2018.

\bibitem{olson2016automating}
R.~S. Olson, R.~J. Urbanowicz, P.~C. Andrews, N.~A. Lavender, J.~H. Moore, {\em
  et~al.}, ``Automating biomedical data science through tree-based pipeline
  optimization,'' in {\em European Conference on the Applications of
  Evolutionary Computation}, pp.~123--137, Springer, 2016.

\bibitem{OpenML2013}
J.~Vanschoren, J.~N. van Rijn, B.~Bischl, and L.~Torgo, ``Openml: Networked
  science in machine learning,'' {\em SIGKDD Explorations}, vol.~15, no.~2,
  pp.~49--60, 2013.

\bibitem{li2016learning}
K.~Li and J.~Malik, ``Learning to optimize,'' {\em arXiv preprint
  arXiv:1606.01885}, 2016.

\end{thebibliography}

\addtolength{\textheight}{-12cm}   


\end{document}